\def\year{2018}\relax
\documentclass[letterpaper]{article} 
\usepackage{aaai182}  
\usepackage{times}  
\usepackage{helvet}  
\usepackage{courier}  
\usepackage{url}  
\usepackage{graphicx}  

\usepackage{amssymb}

\begin{document}
%
\title{AdaDNNs: Adaptive Ensemble of Deep Neural Networks \\ for Scene Text Recognition}
\author{Chun Yang$^\dag$, Xu-Cheng Yin$^{\dag*}$, Zejun Li$^\dag$, Jianwei Wu$^\dag$, \\ 
\textbf{Chunchao Guo}$^\ddag$, \textbf{Hongfa Wang}$^\ddag$, and \textbf{Lei Xiao}$^\ddag$\\
$^\dag$ Department of Computer Science and Technology, University of Science and Technology Beijing, Beijing, China\\
$^\ddag$ TEG, Tencent Co. LTD, Shenzhen, China\\
$^*$ Corresponding author: xuchengyin@ustb.edu.cn\\
}
\maketitle
\begin{abstract}
Recognizing text in the wild is a really challenging task because of complex backgrounds, various illuminations and diverse distortions, even with deep neural networks (convolutional neural networks and recurrent neural networks).
In the end-to-end training procedure for scene text recognition, the outputs of deep neural networks at different iterations are always demonstrated with diversity and complementarity for the target object (text). Here, a simple but effective deep learning method, an adaptive ensemble of deep neural networks (AdaDNNs), is proposed to simply select and adaptively combine classifier components at different iterations from the whole learning system. Furthermore, the ensemble is formulated as a Bayesian framework for classifier weighting and combination.
A variety of experiments on several typical acknowledged benchmarks, i.e., ICDAR Robust Reading Competition (Challenge 1, 2 and 4) datasets, verify the surprised improvement from the baseline DNNs, and the effectiveness of AdaDNNs compared with the recent state-of-the-art methods.
\end{abstract}

\noindent Scene text is widely used as visual indicators for navigation and notification, and text recognition from scene images and videos is one key factor for a variety of practical applications with reading in the wild~\cite{Ye2015,Yin2016,Tian2017}, such as assisting for visually impaired people~\cite{Goto2009,Sanketi2011}, real-time translation~\cite{Shi2005,Fragoso2011}, user navigation~\cite{Minetto2011}, driving assistance systems~\cite{Wu2005}, and autonomous mobile robots~\cite{Letourneau2003}.

Scene text (cropped word) recognition methods can be generally grouped into \textbf{segmentation-based word recognition} and \textbf{holistic word recognition}. Typical segmentation-based approaches over-segment the word image into small segments, combine adjacent segments into candidate characters, classify them using convolutional neural networks (CNNs) or gradient feature-based classifiers, and find an approximately optimal word recognition result~\cite{Bissacco2013,Jaderberg2014}.
Because of complex backgrounds and diverse distortions, character segmentation is another more challenging task. Thereby, holistic word recognition approaches with deep neural networks are more impressive for text reading in the wild. \textbf{Word spotting}, the direct holistic approach, usually calculates a similarity measure between the candidate word image and a query word~\cite{Jaderberg2014c,Gordo2015}. \textbf{Sequence matching}, the indirect holistic approach, recognizes the whole word image by embedding hidden segmentation strategies. For example, Shi et al. constructed an end-to-end training deep neural network for image-based sequence recognition (scene text recognition)~\cite{Shi2017}.

\begin{figure}[htb!]
\centering
  \includegraphics[width=0.35\textwidth]{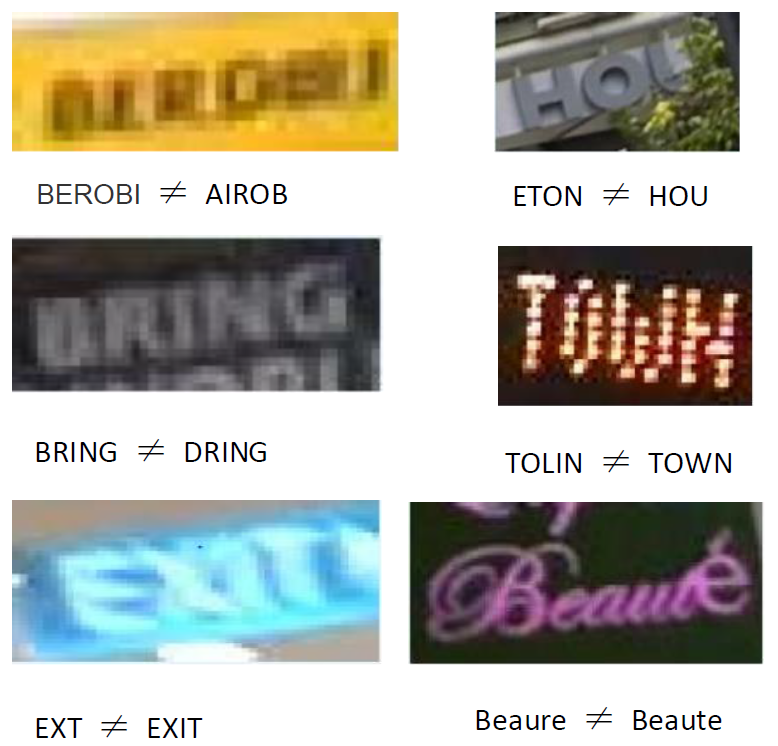}\\
  \caption{\footnotesize{Some challenging examples (from 2015 ICDAR Robust Reading Competition Challenge 4 dataset) of scene text images which are incorrectly recognized by the baseline DNNs (see related descriptions in Experiments). The captions show the recognized text (left) versus the ground truth (right): additional characters, wrong characters and missing characters in target words.}}
  \label{fig:examples}
\end{figure}

However, there are a variety of grand challenges for scene text recognition (see samples in Fig.~\ref{fig:examples}), even with recent deep neural networks (DNNs), where additional characters will be probably identified for text distortions and complex backgrounds, some characters are wrongly recognized for changing illuminations and complex noises, and characters are sometimes missed for low resolutions and diverse distortions.

Stochastic Gradient Descent (SGD)~\cite{Bottou2010} and its variants have become the defacto techniques for optimizing DNNs, where SGD always leads to local minima, even though the popularity of SGD can be attributed to its ability to avoid spurious saddle-points and local minima~\cite{Dauphin2014}. There are a plenty number of (more than million) possible local minima in DNNs~\cite{Kawaguchi2016}, and local minima with flat basins are supposed to generalize better in the learning system~\cite{Keskar2017}.
As a result, although different local minima often have similar error rates, the corresponding neural networks in DNNs tend to make different mistakes. This diversity and complementarity can be exploited via classifier ensemble~\cite{Huang2017}.

There are two major ways for ensemble of deep neural networks. On the one hand, \textbf{different learning systems} with DNNs are first trained independently, and then the final system is a trivial ensemble of these different deep learning architectures via majority voting or averaging. For example, most high profile competitions in ImageNet~\footnote{\url{www.image-net.org}.} and Kaggle~\footnote{\url{www.kaggle.com}.} are won by such ensemble techniques. Because of the huge computation complexity, this ensemble becomes uneconomical and impossible for most researchers in the universities and even in the small companies.
On the other hand, \textbf{one learning system} with DNNs is first trained, and then the final ensemble selects and combines \textbf{neural network components}~\footnote{The neural network component means the resulting DNN of each iteration in the whole training procedure.} in this only one system without incurring any additional training cost. Huang et al. proposed such an ensemble technique, called as Snapshot Ensembling, where a specific optimization strategy is designed to train DNNs and ``model snapshots" (neural network components) in all cycles are combined for the final ensemble in the learning procedure~\cite{Huang2017}. However, how to design the specific and effective optimization algorithms for DNNs is also a challenge.

In this paper, we propose a new and adaptive ensemble of deep neural networks (\textbf{AdaDNNs}) in the most simplest way, i.e., given trained neural networks (of all iterations) from a learned DNNs system~\footnote{Here, the DNNs system can be trained with conventional optimization algorithms~\cite{Bottou2016}, or even with the specific algorithms, e.g., Snapshot Ensembling~\cite{Huang2017}.}, a subset of neural network components are simply selected and adaptively combined to perform the final predictions. And the ensemble is formally formulated as a Bayesian framework for classifier weighting and combination.
We argue that because of the diversity and complementarity in DNNs with SGD, AdaDNNs via ensembling with diversity can improve robust performance of the final learning system. On the same time, because of the high accuracy of components in DNNs, AdaDNNs via combination with accurate neural network components can improve precision performance of the final classification system.
A variety of experiments on several acknowledged benchmarks, i.e., ICDAR Robust Reading Competition (Challenge 1, 2 and 4) datasets, have shown that the \textbf{simple but effective} AdaDNNs improves largely from the baseline DNNs. Moreover, our proposed approach has the top performance compared with the latest state-of-the-art methods.


\section{Related Work}

Recognizing text in scene videos attracts more and more interests in the fields of document analysis and recognition, computer vision, and machine learning. The existing methods for scene text (cropped word) recognition can be grouped into segmentation-based word recognition and holistic word recognition. In general, segmentation-based word recognition methods integrate character segmentation and character recognition with language priors using optimization techniques, such as Markov models~\cite{Weinman2014} and CRFs~\cite{Mishra2012,Shi2013}.
In recent years, the mainstream segmentation-based word recognition techniques usually over-segment the word image into small segments, combine adjacent segments into candidate characters and classify them using CNNs or gradient feature-based classifiers, and find an approximately optimal word recognition result using beam search~\cite{Bissacco2013}, Hidden Markov Models~\cite{Alsharif2014}, or dynamic programming~\cite{Jaderberg2014}.

Word spotting~\cite{Manmatha1996}, a direct holistic word recognition approach, is to identify specific words in scene images without character segmentation, given a lexicon of words~\cite{Wang2010}.
Word spotting methods usually calculate a similarity measure between the candidate word image and a query word. Impressively, some recent methods design a proper CNN architecture and train CNNs directly on the holistic word images~\cite{Jaderberg2014b,Jaderberg2014c}, or use label embedding techniques to enrich relations between word images and text strings~\cite{Almazan2014,Gordo2015}. Sequence matching, an indirect holistic word recognition approach, recognizes the whole word image by embedding hidden segmentation strategies. Shi et al. constructed an end-to-end train deep neural network for image-based sequence recognition (scene text recognition), where a convolutional recurrent neural networks framework (CRNN) is designed and utilized~\cite{Shi2017}. In this paper, a similar CRNN architecture is used in AdaDNNs for recognizing scene text sequently and holistically.


Classifier ensemble can be mainly divided into two categories. The first one aims at learning multiple classifiers at the feature level, where multiple classifiers are trained and combined in the learning process, e.g., Boosting~\cite{Freund97}, Bagging~\cite{Breiman96}, and Rotation Forest~\cite{Rodriguez06}. The second tries to combine classifiers at the output level, where the results of multiple available classifiers are combined to solve the targeted problem, e.g., multiple classifier systems (classifier combination)~\cite{Zhou12,Yin2014inffus}.
AdaDNNs in this paper follows the second one. Namely, given multiple classifiers (neural network components sequently learned in DNNs), AdaDNNs is constructed by combining intelligently these component classifiers within a Bayesian-based formulation framework.

\section{Adaptive Ensemble of Deep Neural Networks}
As we have known, both SGD and batch optimization can lead to different local minima in DNNs, and neural network components are always with diversity and complementarity. Conventionally, there are tens of thousands of iterations and also neural network components in the learning system of DNNs. Considering the acceptable computation complexity in the testing procedure, one thing is to quickly select a small subset of neural network components in different training iterations. At the same time, considering the high accuracy requirement, another thing is to adaptively combine this subset of neural network components and construct a final classification system.
In the following, the unified framework of AdaDNNs is first formulated. Next, the detail procedure of AdaDNNs is then described.

\subsection{Unified Framework}
To formulate the ensemble decision, the individual classifier decisions can be combine by majority voting, which sums the votes for each class and selects the class that receives most of the votes.  While the majority voting is the most popular combination rule, a major limitation of majority voting is that only the decision of each model is taken into account without considering the distribution of decisions.

In particular, all the possible models in the hypothesis space could be exploited by considering their individual decisions and the correlations with other hypotheses. Here, we use a Bayesian-based framework to combine classifiers. Given a sample $x$ and a set $H$ of independent classifiers, the probability of label $y$ can be estimated by a Bayesian Model (BM) as
\begin{equation}\label{eq:1}
 P(y|H,x) = \sum_{h_i \in H} P(y|h_i,x)P(h_i|x)
\end{equation}
where $P(y|h_i,x)$ is the distribution of describing the correlation between decision $y$ and $h_i(x)$, and $P(h_i|x)$ denotes the posterior probability of model $h_i$. The posterior $P(h_i|x)$ can be computed as
\begin{equation}\label{eq:2}
 P(h_i|x) = \frac{P(D|h_i)P(h_i)}{\sum_{h_i \in H}P(D|h_i)P(h_i)}
\end{equation}
where $P(h_i)$ is the prior probability of classifier $h_i$ and $P(D|h_i)$ is the model likelihood on the training set $D$.
Here, $P(h_i)$ and $\sum_{h_i \in H}P(D|h_i)P(h_i)$ are assumed to be a constant in Eq.~\ref{eq:2}.
Therefore, BM assigns the optimal label $y$ to $y^*$ according to the following decision rule, i.e.,
\begin{equation}\label{eq:3}
\begin{array}{rl}
 P(y^*) &= argmax_{y}P(y|H,x) \\
        &= argmax_{y}\sum_{h_i\in H} P(y|h_i,x)P(h_i|x) \\
        &= argmax_{y}\sum_{h_i\in H} P(y|h_i,x)P(D|h_i) \\
        &= argmax_{y}2\sum_{h_i\in H} P(y|h_i,x)P(D|h_i) - P(D) \\
        &= argmax_{y}\sum_{h_i\in H} (2P(y|h_i,x)-1)P(D|h_i) \\
        &= argmax_{y}\sum_{h_i\in H} W(y,h_i(x))P(D|h_i) \\
\end{array}
\end{equation}
where $W(y,h_i(x))$ is a function of $y$ and $h_i(x)$. By multiplying the scaling factor $\lambda > 0$, $W(y,h_i(x))$ can have a different range in $\mathbb{R}$.

There are two key issues for optimizing Eq.~\ref{eq:3}. The first one is the calculation of $W(y,h_i(x))$. As mentioned above, $P(y|h_i,x)$ is the distribution of describing the correlation between decision $y$ and $h_i(x)$. Thus, $W(y, h_i(x))$ can be derived from $y$, $h_i(x)$ and the distance between $y$ and $h_i(x)$. Here, $W(y,h_i(x))$ is assumed to be computed as
\begin{equation}\label{eq:4}
 W(y, h_i(x)) = I(y=h_i(x)) + U(y)*V(y,h_i(x))
\end{equation}
where both $U(*)$ and $V(*,\bullet)$ are functions. $I(y=h_i(x))$ returns $1$ when $y=h_i(x)$; otherwise, $I(y=h_i(x)) = 0$.

\begin{figure}[htb!]
\centering
  \includegraphics[width=0.35\textwidth]{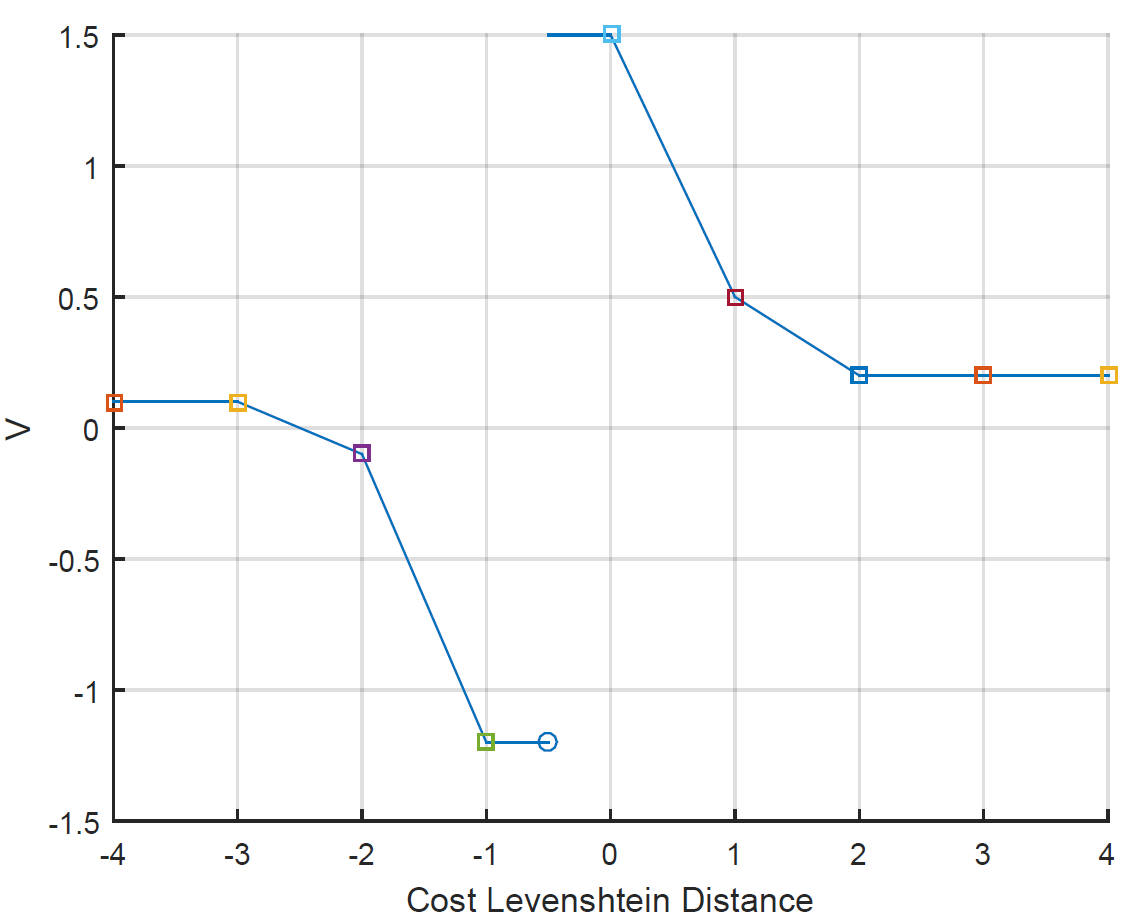}\\
  \caption{\footnotesize{An example of $F$ of describing the relationship between $V$ (in Y-axis) and  the Cost Levenshtein Distance (in X-axis).}}
  \label{fig:cld}
\end{figure}

For the scene text recognition task, on the one hand, with a given dictionary~\footnote{In our experiments, a $90k$ word dictionary from~\cite{Jaderberg2014c} is used as the given dictionary.}, $U(y)$ can be calculated as
\begin{equation}\label{eq:5}
U(y) =  \{
      \begin{array}{ll}
      1                      &y \in Dict\\
      0                      &y \notin Dict\\
      \end{array}
\end{equation}
On the other hand, the correlation between $y$ and $h_i(x)$ can be assumed by the function $V$ of Cost Levenshtein Distance (CLD). In the  traditional Levenshtein Distance, the cost of any two different characters is always $1$. However, in spelling correction, the cost of two characters with similar shape tends to have a smaller distance.
In this paper, we statistics the frequencies of different character pairs at the same location from the label and the hypothesis on the validation set (bootstrapped from the training set in Experiments), and calculate the cost of two different characters ($a$ and $b$) as
\begin{equation}\label{eq:6}
 cost(a,b) = 1-P(a|b)
\end{equation}
Note that if both $y$ and $h_i(x)$ are from the given dictionary, then they will have a competitive relationship with each other. Thus, $V(y,h_i(x))$ can be calculated with
\begin{equation}\label{eq:5}
 V(y,h_i(x)) = \{
    \begin{array}{rl}
      F(-CLD(y,h_i(x))) & h_i(x) \in Dict \\
      F(CLD(y,h_i(x))) & h_i(x) \notin Dict \\
    \end{array}
\end{equation}
where $F$ is a function of the CLD between $y$ and $h_i(x)$.
By a heuristic approach, the values of $F$ can be empirically assigned at the multiple integral points, and the values at other points can be calculated by the piecewise linear interpolation. An example of $F$ is shown in Fig.~\ref{fig:cld}. In general, $F$ has a small range, e.g., $[1.5, 1.5]$ in Fig.~\ref{fig:1}. So, the obtained weights from Eq.~\ref{eq:4} are convenient for linear combination of classifiers.

The second issue is about generating voting candidates (more probable labels of the hypotheses). Obviously, the ground truth doesn't always appear in the decisions made by $H$. It is necessary to find an effective way to generate good candidates from all the decisions, i.e., to find a more probable label $y_i(x)$ from the existed initial label $y_i^{0}(x)$ of hypothesis $h_i$. Generally speaking, a good candidate means it has a small edit distance with most of the hypotheses.
Following this idea, we propose an algorithm to semantically generate voting candidates (see Algorithm $1$).
\begin{center}
\small
\begin{tabular}{|l|}
\hline
  Algorithm $1$: Generating Voting Candidates.\\
\hline
  \textbf{Input:}\\
  ~~~~~~~$H=\{h_1,h_2,...,h_L\}$: the base classifier set, $|H|=L$.\\
  ~~~~~~~$Y_0$: the initial decisions made by $H$. \\
  ~~~~~~~$ED$: the measurement function of the pairwise distance.\\
  ~~~~~~~$\theta$: the upper bound of the distance between the candidate \\
  ~~~~~~~~~~~and the hypothesis. \\
  \textbf{Output:} \\
  ~~~~~~~$Y$: the voting candidate set.\\
  \textbf{Parameter:}\\
  ~~~~~~~$H^{\star}$: a subset of H, $\forall h^{\star}_i,h^{\star}_j \in H^{\star}, ED(h^{\star}_i,h^{\star}_j) \leq 2\theta$. \\
  \textbf{Procedure:}\\
  ~1:~~~~~~$Y = \emptyset$. \\
  ~2:~~\textbf{For} each $H^{\star} \subset H$;\\
  ~3:~~~~\textbf{For} each $y \in Y_0$: \\
  ~4:~~~~~~\textbf{If} $max_{h^{\star}_i \in H^{\star}} ED(y,h^{\star}(x)) \leq \theta$: \\
  ~5:~~~~~~~~$Y = Y \cup \{y\}$. \\
  ~6:~~~~\textbf{End}\\
  ~7:~~\textbf{End}\\
  \hline
\end{tabular}
\end{center}

In Algorithm 1, the searching process of $H^{\star}$ is an implicit computational way for $P(D|h_i)$.
In our experiments, a special simple case of algorithm $1$ is used, where during the voting candidates generation process, $Y_0$ is initialized only by $H$, the upper bound is set from $\theta$ to $\inf$, and $P(D|h_i)$ is assumed to be a constant.

\subsection{AdaDNNs Algorithm}
Within the above framework, the procedure of AdaDNNs for scene text recognition includes three major steps, i.e., base classifiers generation, classifier combination, and ensemble pruning.

\subsubsection{Base Classifiers Generation}
Ensembles work best if the base models have high accuracy and do not overlap in the set of examples they misclassify.
Deep neural networks (DNNs) are naturally used as a base classifier generator for ensembles.
On the one hand, DNNs have dramatically improved the state-of-the-art in many domains, such as speech recognition, visual object recognition and object detection, by being composed of multiple processing layers to learn representations of data with multiple levels of abstraction.
On the other hand, during the training phase of one individual deep neural network, two snapshots with different minibatch orderings will converge to different solutions. Those snapshots often have the similar error rates, but make different mistakes.
This diversity can be exploited by ensembling, in which multiple snapnots are average sampling and then combined with majority voting.

Focusing on scene text recognition, the CRNN model~\cite{Shi2017} is used to generate base classifiers (neural network components) as our text recognizer. CRNN uses CTC~\cite{GravesFGS06} as its output layer, which estimates the sequence probability conditioned on the input image, i.e. $P(h|x)$, where $x$ is the input image and $h$ represents a character sequence.

\subsubsection{Classifier Combination for AdaDNNs}
The core of AdaDNNs is to calculate $y^*$ (by Eq.~\ref{eq:3}), i.e., the calculation of $F$, which is a function of distance between $y$ and $h_i(x)$.
Here, $F$ is represented by the set of values at the multiple integral points.
These values are assigned with the highest recognition rate on the validation set.
The detail procedure of the AdaDNNs ensemble is shown in Algorithm $2$.
\begin{center}
\small
\begin{tabular}{|l|}
\hline
  Algorithm $2$: AdaDNNs (classifier combination).\\
\hline
  \textbf{Input:}\\
  ~~~~~~~$H=\{h_1,h_2,...,h_L\}$: the base classifier set, $|H|=L$.\\
  ~~~~~~~$Dict$: the given dictionary. \\
  ~~~~~~~$F$: a function of distance between $y$ and $h_i(x)$. \\
  \textbf{Parameter:}\\
  ~~~~~~~$Y$: the voting candidates set generated by Algorithm $1$. \\
  \textbf{Output:} \\
  ~~~~~~~$y^*$: the label of prediction.\\
  \textbf{Procedure:}\\
  ~1:~~~~~~Initialize $Y$ by $H$ and $Dict$. \\
  ~3:~~\textbf{For} $y \in Y$: \\
  ~4:~~~~~~Calculate $P(y|H,x)$ through Eq. \ref{eq:2}.\\
  ~5:~~\textbf{End} \\
  ~6:~~~~~~Calculate $y^*$ through Eq. \ref{eq:3}.\\
  \hline
\end{tabular}
\end{center}

\subsubsection{AdaDNNs Pruning}
In classifier ensemble, pruning can generally improve the ensemble performance.
Here, we use Genetic Algorithm (GA) to pruning the ensemble.
 GA is a meta heuristic inspired by the process of natural selection that belongs to the larger class of evolutionary algorithms. GAs are commonly used to generate high-quality solutions for optimization and search problems by relying on bio-inspired operators such as mutation, crossover and selection.

In AdaDNNs pruning, firstly, a population of binary weight vectors is randomly generated, where $1$ means the classifier is remained. Secondly, the population is iteratively evolve where the fitness of a vector $w$ is measured on the validation set $V$, i.e., $f(w)=R^V_w$ ($R$ stands for the recognition rate). Finally, the ensemble is correspondingly pruned by the evolved best weight vector $w^*$.

\section{Experiments}

To evaluate the effectiveness of the proposed AdaDNNs method, a variety of experiments for text (cropped word) recognition are conducted on acknowledged benchmark datasets.
We first focused on the most challenging task, i.e., incidental scene text recognition (ICDAR Robust Reading Competition Challenge 4), trained our AdaDNNs learning system (on both the synthetic dataset from~\cite{Jaderberg2014c} and the training set of Challenge 4), and performed comparative experiments.
Then, we also conducted experiments of this learned AdaDNNs on other text recognition tasks, i.e., focused scene text recognition and born-digital text recognition (ICDAR Robust Reading Competition Challenge 1 and 2), and checked the generalization of AdaDNNs.
Here, the baseline DNNs model, CRNN, is same to the one in~\cite{Shi2017}. The official metrics in ICDAR 2011/2013/2015 Robust Reading Competition~\cite{icdar2011,2013Challenge,2015Challenge} are used.

\subsection{Experiments with Incidental Scene Text Recognition}

The ICDAR 2015 Robust Reading Competition Challenge 4 database~\cite{2015Challenge} is a widely used and highly competitive benchmark database for scene text recognition within complex situations in the recent $3$ years. The public dataset includes a training set of $1,000$ images and a test set of $500$, with more than $10,000$ annotated text regions (cropped words). Because of complex backgrounds, various illuminations and diverse distortions, this incidental scene text recognition topic is a very challenging task. In our experiments, a variety of methods are conducted and compared, i.e., the baseline DNNs, AdaDNNs, AdaDNNs pruning, the winning participation method in the official competition (marked as \textbf{bold words}),
and the latest top submissions of the Robust Reading Competition (RRC) website~\footnote{\url{http://rrc.cvc.uab.es}.} in 2017 (marked as \emph{italic words}).

\begin{table}[htp!]
     \centering\caption{\label{tab:ch4} Comparative results on 2015 ICDAR Challenge 4 dataset (incidental scene text recognition), where the comparative results are from the RRC website.}
    \tiny
    \begin{tabular}{|r|r|rrrr|}
    \hline
Date	 & Method	 & T.E.D	 & C.R.W	 & T.E.D.	 & C.R.W.	\\
         &           &           & (\%)      & (upper)           & (upper)          \\\hline
\emph{2017/7/6}	 & \emph{Baidu IDL v3}	 & 211.59	 & \textbf{80.02}	 & 171.15	 & 82.33	\\
\emph{2017/7/6}	 & \emph{HIK\_OCR\_v3}	 & \textbf{191.25}	 & 78.29	 & 158.84	 & 80.12	\\
\emph{2017/6/29}	 & \emph{HKU-VisionLab}	 & 258.59	 & 72.03	 & 212.17	 & 74.19	\\ \hline
\textbf{2015/4/1}	 & \textbf{MAPS}	 & 1,128.01	 & 32.93	 & 1,068.72	 & 33.90	\\ \hline
-	 & Baseline DNNs	 & 384.76	 & 60.18	 & 303.77	 & 64.90	\\
-	 & AdaDNNs	 & 251.98	 & 76.31	 & 185.36	 & 80.55	\\
-	 & AdaDNNs Pruning	 & 224.7	 & 79.78	 & \textbf{147.11}	 & \textbf{84.21}	\\\hline
\end{tabular}
\end{table}

As can be seen from Table~\ref{tab:ch4}, our proposed AdaDNNs is much better than the baseline DNNs. For example, for the measure of ``C.R.W (upper)", AdaDNNs has a surprised improvement, i.e., from $64.90\%$ to $80.55\%$. That is to say, the adaptive ensemble of DNNs in a simple but effective strategy can largely improved the performance from the original baseline DNNs.
Moreover, compared with the latest top submissions (e.g., ``Baidu IDL v3" and ``HIK\_OCR\_v3"), our method, AdaDNNs Pruning, has the best performance with ``C.R.W (upper)", i.e., $84.21\%$.

\begin{figure}[htb!]
\centering
  \includegraphics[width=0.46\textwidth]{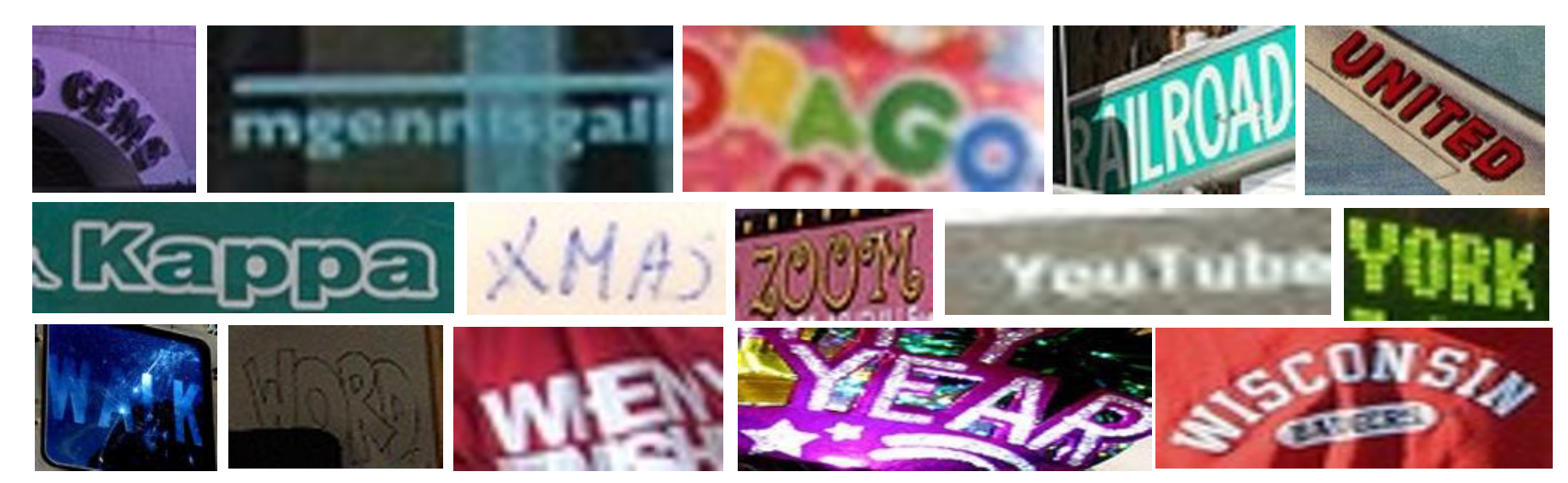}\\
  \caption{\footnotesize{Challenging samples of scene text from COCO-text which are correctly recognized (with C.R.W. upper) by AdaDNNs: ``GEMS", ``mgennisgal", ``RGAO", ``RAILROAD", ``UNITED", ``Kappa", ``XMAS", ``ZOOM", ``YouTube", ``YORK", ``WALK", ``WPRD", ``WHEN", ``YEAR", and ``WISCONSIN".}}
  \label{fig:coco-text}
\end{figure}

We also perform experiments on the COCO-text dataset~\cite{Veit2016}, a similar challenging but large-scale incidental scene text dataset. Images in this dataset are from the MS COCO dataset that contain text ($63,686$ images with $173,589$ text regions). ICDAR2017 Robust Reading Challenge on COCO-Text is holding and will be released in ICDAR 2017. So, the comparative results of AdaDNNs, AdaDNNs Pruning and the baseline DNNs are only on the validation set; they are $58.08\%$, $66.07\%$, and $66.27\%$, respectively. Some scene text recognition samples for COCO-text are shown in Fig.~\ref{fig:coco-text}.

\subsection{Experiments with Focused Scene Text Recognition and Born-Digital Text Recognition}

In order to investigate the generalization of AdaDNNs, we directly use the trained AdaDNNs system above (for 2015 ICDAR Challenge 4), and perform experiments on 2013 ICDAR Challenge 2 (cropped word recognition) dataset. The Challenge 2 dataset contains $1,015$ ground truths cropped word images.
In our experiments, a variety of methods are conducted and compared, i.e., the baseline DNNs, AdaDNNs, AdaDNNs pruning, the winning participation method in the official competition (marked as \textbf{bold words}), the top three results in published papers,
and the latest top submissions of the RRC website in 2017 (marked as \emph{italic words}).

\begin{table}[htp!]
     \centering\caption{\label{tab:ch2} Comparative results on 2013 ICDAR Challenge 2 dataset (focused scene text recognition), where the comparative results without publications are from the RRC website.}
    \tiny
    \begin{tabular}{|r|r|rrrr|}
    \hline
Date	 & Method	 & T.E.D	 & C.R.W	 & T.E.D.	 & C.R.W.	\\
         &           &           & (\%)      & (upper)           & (upper)          \\\hline
\emph{2017/8/14}	 & \emph{TencentAILab}	 & \textbf{42}	 & \textbf{95.07}	 & \textbf{39.35}	 & \textbf{95.34}	\\
\emph{2017/7/28}	 & \emph{Tencent Youtu}	 & 48.12	 & 92.42	 & 40.37	 & 93.42	\\
\emph{2017/2/24}	 & \emph{HIK\_OCR}	 & 64.95	 & 90.78	 & 42.31	 & 93.33	\\ \hline
2016	 & CNN~\cite{Jaderberg2014c}	 & --	 & --	 & --	 & 90.8	\\
2016	 & RARE~\cite{Shi2016}	 & --	 & --	 & --	 & 88.6	\\
2017	 & CRNN~\cite{Shi2017}	 & --	 & --	 & --	 & 89.6	\\ \hline
\textbf{2013/4/6}	 & \textbf{PhotoOCR}	 & 122.75	 & 82.83	 & 109.9	 & 85.30	\\ \hline
-	 & Baseline DNNs	 & 306.43	 & 75.34	 & 282.35	 & 78.63	\\
-	 & AdaDNNs	 & 193.51	 & 83.20	 & 170.13	 & 86.67	\\
-	 & AdaDNNs Pruning	 & 182.7	 & 85.21	 & 164.38	 & 88.13	\\\hline
\end{tabular}
\end{table}

Similarly, AdaDNNs is much better than the baseline DNNs, e.g., the measure of ``C.R.W (upper)" increases from $78.63\%$ to $86.67\%$.
Surprisedly, only trained for another task (Challenge 4), the AdaDNNs (AdaDNNs Pruning) has a competitive performance on a new dataset   (Challenge 2 dataset), compared with the recent published methods (e.g., CRNN~\cite{Shi2017}), and even with the latest submission results.

Apart from the above experiments on text recognition from scene images (ICDAR Robust Reading Competition Challenge 2 and 4), we also directly perform the learned AdaDNNs on the born-digital images track (Challenge 1). Though born-digital images are not scene images, they have similar challenging issues for text recognition, e.g., complex backgrounds, low resolution and various colors. 
We also compare AdaDNNs (AdaDNNs Pruning) with the baseline DNNs, the winning participation method in the official competition (marked as \textbf{bold words}), and the latest top submissions of the RRC website (marked as \emph{italic words}). The similar conclusions are drawn. Firstly, AdaDNNs improves largely compared with the baseline DNNs (from $84.50$ to $92.22$ for ``C.R.W (upper)"). Secondly, AdaDNNs has a comparative performance with the latest submission results (e.g., ``Dahua\_OCR\_v1" with $92.49\%$ in 2017/9/1).

\begin{table}[htp!]
     \centering\caption{\label{tab:ch1} Comparative results on ICDAR Challenge 1 dataset (born-digital text recognition), where the comparative results are from the RRC website.}
    \tiny
    \begin{tabular}{|r|r|rrrr|}
    \hline
Date	 & Method	 & T.E.D	 & C.R.W	 & T.E.D.	 & C.R.W.	\\
         &           &           & (\%)      & (upper)           & (upper)          \\\hline
\emph{2017/8/22}	 & \emph{Tecent Youtu}	 & \textbf{17.51}	 & \textbf{96.80}	 & 13.67	 & \textbf{97.29}	\\
\emph{2017/7/21}	 & \emph{TecentAILab}	 & 18.77	 & 96.18	 & \textbf{12.91}	 & 97.22	\\
\emph{2017/9/1}	 & \emph{Dahua\_OCR\_v1}	 & 57.47	 & 91.31	 & 42.87	 & 92.49	\\\hline
\textbf{2013/4/6}	 & \textbf{PhotoOCR}	 & 103.41	 & 82.21	 & 87.19	 & 85.41	\\\hline
-	 & Baseline DNNs	 & 87.7	 & 82.42	 & 72.32	 & 84.50	\\
-	 & AdaDNNs	 & 55.44	 & 89.58	 & 39.61	 & 92.22	\\
-	 & AdaDNNs Pruning	 & 55.64	 & 89.53	 & 39.81	 & 92.17	\\\hline
\end{tabular}
\end{table}

We fully believe that if AdaDNNs (AdaDNNs Pruning) performs re-training on ICDAR Challenge 1 and Challenge 2 datasets, the performance will correspondingly be improved and obtain a more impressive results compared with the latest submission systems. This is also a near issue for our future work.

\section{Conclusion and Discussion}

A variety of DNNs based methods have been proposed and are still being investigated in the literature for scene text recognition because of the grand challenges, e.g., complex backgrounds, various illuminations and diverse distortions. In order to fully take advantage of the complementary diversity and the high accuracy of neural network components in DNNs, an adaptive ensemble of deep neural networks (AdaDNNs) is proposed to simply select and adaptively combine neural networks in the whole training procedure. Comparative experiments of scene text (cropped word) recognition showed that AdaDNNs achieves a remarkable increase in the final performance (more than $10\%$) compared with the baseline DNNs.

Note that the DNNs methods have dramatically improved the state-of-the-art in object detection, object recognition, speech recognition and many other domains. 
Consequently, a near future issue is to evaluate the efficacy of AdaDNNs with state-of-the-art DNNs on object recognition and speech recognition. For example, experiments for object detection and recognition of AdaDNNs with Snapshot Ensembling~\cite{Huang2017}, ResNet~\cite{He2016}, and DenseNet~\cite{Huang2017b} can be performed and compared in the next step.

\bibliography{CNNEnsemble}
\bibliographystyle{aaai}

\end{document}